\documentclass[10pt,twocolumn,letterpaper]{article}

\usepackage{cvpr}
\usepackage{times}
\usepackage{epsfig}
\usepackage{graphicx}
\usepackage{amsmath}
\usepackage{amssymb}
\usepackage{makecell}
\usepackage{multirow}
\usepackage{booktabs}

\usepackage{algorithm}
\usepackage{algorithmic} %format of the algorithm
\usepackage{dsfont}
\usepackage{amsfonts}

\usepackage{pifont}% http://ctan.org/pkg/pifont
\usepackage[pagebackref=true,breaklinks=true,letterpaper=true,colorlinks,bookmarks=false]{hyperref}

\cvprfinalcopy % *** Uncomment this line for the final submission

\def\cvprPaperID{0006} % *** Enter the CVPR Paper ID here

\ifcvprfinal\fi

\begin{document}

%%%%%%%%% TITLE
\title{State-aware Re-identification Feature for Multi-target Multi-camera Tracking}

\author{Peng Li$^{*1,3}$, Jiabin Zhang$^{*2}$, Zheng Zhu$^{*2}$, Yanwei Li$^{2}$, Lu Jiang$^{3}$, Guan Huang$^{3}$\\
$^{1}$Beijing University of Posts and Telecommunications, Beijing, China\\
$^{2}$Institute of Automation, Chinese Academy of Sciences, Beijing, China\\
$^{3}$Horizon Robotics, Beijing, China\\
{\tt\small qqlipeng@bupt.edu.cn zhengzhu@ieee.org}
\\
{\tt\small \{zhangjiabin2016,liyanwei2017\}@ia.ac.cn}
\\
{\tt\small \{lu.jiang,guan.huang\}@horizon.ai}
}

\maketitle
\thispagestyle{empty}

%%%%%%%%% ABSTRACT
\begin{abstract}
   Multi-target Multi-camera Tracking (MTMCT) aims to extract the trajectories from videos captured by a set of cameras. Recently, the tracking performance of MTMCT is significantly enhanced with the employment of re-identification (Re-ID) model. However, the appearance feature usually becomes unreliable due to the occlusion and orientation variance of the targets. Directly applying Re-ID model in MTMCT will encounter the problem of identity switches (IDS) and tracklet fragment caused by occlusion. To solve these problems, we propose a novel tracking framework in this paper. In this framework, the occlusion status and orientation information are utilized in Re-ID model with human pose information considered. In addition, the tracklet association using the proposed fused tracking feature is adopted to handle the fragment problem. The proposed tracker achieves 81.3\% IDF1 on the multiple-camera hard sequence, which outperforms all other reference methods by a large margin.
\end{abstract}
\footnotetext{*The first three authors contributed equally to this work.}
%%%%%%%%% BODY TEXT
\section{Introduction}
Multi-target Multi-camera Tracking (MTMCT) is a significant problem in computer vision and is particularly useful for public security and video understanding \cite{zhu2018endto,zhu2018two}. MTMCT aims to track multiple targets across multiple cameras, which is different from the multi-object tracking (MOT) in single camera \cite{wang2013intelligent}. Camera network has a broader view than a single camera and has a broader foreground of applications. However, in addition to facing the same challenges of occlusion, pose variance and background clutter with MOT, MTMCT also faces some specific challenges like the blind area among cameras, change of viewpoint and illumination variance.

Feature representation, occlusion handling, and inference are critical components for both MOT and MTMCT. In this paper, we concentrate on the first two components. Appearance feature is significant to maintain the identity of the tracked target, and many works \cite{chu2017online,zhu2018online,feng2019multi} have exploited the reliable appearance model. In specific, color histogram \cite{mitzel2011real,izadinia20122} and HOG \cite{kuo2010multi,choi2012unified} are well studied and utilized in previous works. However, color histogram and HOG are not robust to the occlusion, and they can not handle the appearance variance well. Recently, re-identification (Re-ID) model is widely adopted as a discriminative appearance descriptor. Besides, person Re-ID is closely related to MTMCT, so the high-quality Re-ID feature always leads to a high tracking performance, which has been proved in \cite{ristani2018features}. However, Re-ID training data is usually labeled manually, and highly occluded samples are always discarded from the training data. Therefore, using the Re-ID feature directly with the low-quality detector in a crowded scenario always leads to inferior performance. % Many methods \cite{chu2017online,zhu2018online,feng2019multi} are proposed to address the problems caused by low-quality detector or occlusion.

Occlusion is perhaps the most critical challenge in MOT.
It is a primary cause for ID switches or fragmentation of
trajectories \cite{luo2014multiple}. Directly extracting the feature from the detection region where the target is highly occluded is unreasonable. Therefore, occlusion awareness is crucial for feature extraction. If occlusion status is obtained, only the stable feature can be retained and the occluded feature can be discarded. Besides, orientation has a significant influence on target appearance, which is neglected by most Re-ID models. In \cite{jiang2018online}, the orientation cue is fully exploited. Specifically, the orientation aware loss is proposed to handle the inconsistent problem by orientation variance. In this work, an orientation-aware feature is used to deal with the inconsistent problem.

In the training process of Re-ID task, one identity contains a limited number of instances, but the length of a trajectory in the tracking scene is not limited. Methods like \cite{feng2019multi} and \cite{wojke2017simple} adopt latest Re-ID feature to represent the appearance feature of trajectory. Besides, \cite{zhang2017multi} utilizes the averaged Re-ID feature as a stable representation, which is a common way to use Re-ID feature. However, appearance varies primarily due to the change of background, pose variance, orientation and viewpoints change. Most existing Re-ID models can not handle these problems. Therefore, post-processing on the Re-ID features is necessary for tracking.

Online trackers \cite{bae2014robust,yu2016poi,chu2017online} build the trajectories with the frame by frame association and they usually only consider the relationship between the trajectories and detections. However, the detection result of the occluded target is always inaccurate, and online trackers may produce many fragmented trajectories in this situation. Unlike online trackers, offline trackers like \cite{wang2018exploit} generate the short tracklets at first and link tracklets to get the final trajectories. In addition, offline trackers usually achieve better performance on account of that they can obtain the entire sequence beforehand, and tracklets contain more information than detections when associating. In this work, the tracklet association is adopted to handle the tracklet fragment.

We focus on handling the above issues. The state-aware Re-ID feature is proposed which focuses on appearance representation with extra human pose information. Specifically, human pose information is utilized to estimate the target state which includes the occlusion status and orientation for making better use of the Re-ID feature. Fused tracking feature is designed as the appearance representation of the tracklet for the stable and accurate association in tracking. A distance matrix with the fused tracking feature is proposed for data association.
To handle the fragment of trajectory, the tracklet association is proposed, which includes tracklet rectifying and tracklet clustering. At last, the effectiveness of our framework is verified in the experiment.

The contributions of the paper are listed as follows:

First, human pose information is adopted to infer the target state including the occlusion status and orientation. The novel fused tracking feature is proposed to make the tracking procedure more robust in the crowed scene.

% 具体说下
Second, a redesigned distance matrix on data association is proposed to effectively address the occlusion problem. Besides, a novel tracklet association method is designed to deal with the tracklet fragment problem.

Third, our MTMCT tracker with the state-aware Re-ID feature achieves a new state-of-the-art result on Duke MTMCT benchmark \cite{duke}. Specifically, the submitted result achieves 81.3 \% IDF1 on the multiple-camera hard sequence.

% contribution
\vspace{-0.3cm}
\section{Related Works}
In this section, we introduce previous works on single camera tracking, multiple camera tracking and appearance feature.
\subsection{Single camera tracking}
With the development of object detection, data association is widely adopted in a tracking-by-detection framework. Many methods attempt to adopt global optimization as offline methods like \cite{Wen2018Learning,sheng2018iterative,shen2018tracklet,tang2016multi,tang2017multiple,dehghan2015gmmcp,tang2015subgraph,babaee2018multiple,wang2018exploit}. On the other hand, some methods try to solve data association in an online manner like \cite{solera2015learning,yu2016poi,gao2018osmo,xiang2015learning,gan2018online,sun2018deep,wojke2017simple,chu2017online}. Offline methods usually generate short but accurate tracklets then construct a graph on them and search optimum solution on the graph to get final trajectories. In \cite{dehghan2015gmmcp}, Dehghan \emph{et al.} consider all pairwise relationship between targets and models the data association as a Generalized Maximum Multi Clique
problem (GMMCP). In \cite{tang2015subgraph}, Tang \emph{et al.} formulate the data association as a minimum cost subgraph multicut problem. The graph can link the detections across space and time to handle the long term occlusion. % In \cite{tang2016multi}, Tang \emph{et al.} extend the previous work in \cite{tang2015subgraph} by introducing the deep matching \cite{weinzaepfel2013deepflow}, which is robust to partial occlusion and camera motion. In \cite{berclaz2011multiple}, Berclaz \emph{et al.} utilizes constrained flow optimization on data association and solve it using the k-shortest paths algorithm.

On the other hand, online methods usually match the maintained tracklets with the detections frame by frame. In \cite{xiang2015learning}, Xiang \emph{et al.} use the decision making in Markov decision processes to formulate the online MOT, and reinforcement learning is adopted to learn the similarity function. In \cite{yu2016poi}, Yu \emph{et al.} propose a simple tracking pipeline with high-quality detection and deep learning based appearance feature, which leads to an excellent tracking result. %In \cite{wojke2017simple}, Wojke \emph{et al.} extend the previous work in \cite{bewley2016simple}.
The tracking-by-detection framework heavily depends on the detection quality. With the development of single object tracking (SOT) \cite{kristan2018sixth,zhu2017uct,zhu2018end,li2018high,zhu2018distractor,bai2018multi}, some MOT methods \cite{chu2017online,zhu2018online} with SOT are proposed to handle the problems caused by the inaccurate detections. In \cite{chu2017online}, Chu \emph{et al.} introduce the SOT in MOT framework, and spatial-temporal attention mechanism (STAM) is adopted to handle the drift problems caused by SOT. In \cite{zhu2018online}, Zhu \emph{et al.} propose an extended Efficient Convolution Operators (ECO) \cite{danelljan2017eco} with cost-sensitive tracking loss and introduce Dual Matching Attention Networks (DMAN) with both spatial and temporal attention mechanisms for data association.

\subsection{Multiple cameras tracking}
Multi-target Multi-camera Tracking is a challenging task due to the illumination variance, change of viewpoints and the blind area among cameras. Methods like \cite{kuo2010inter,narayan2017person,wu2017track,gilbert2006tracking,calderara2008bayesian,makris2004bridging} aim to model the relationship among cameras including illumination changes, travel time and entry/exit rates across pairs of cameras. Illumination always varies largely on different viewpoints, so the brightness transfer function (BTF) from a given camera to another camera is estimated to model the illumination changes. \cite{javed2005appearance} finds that all BTFs lie in a low dimensional subspace, and demonstrates that subspace can be used to compute appearance similarity. \cite{prosser2008multi} employs a Cumulative Brightness Transfer Function (CBTF) for mapping color among cameras located at different physical sites. However, the above methods only address the appearance information but ignore the spatial relationship among cameras. To solve this problem, \cite{javed2008modeling} uses kernel density estimation to infer the inter-camera relationships in the form of the multivariate probability density of space-time variables, then integrates spatial cue and appearance cue with the maximum likelihood estimation framework.

In addition, numerous graph-based models \cite{hofmann2013hypergraphs,berclaz2011multiple,tesfaye2017multi,liu2017multi,wen2017multi,yoon2018multiple, wan2013distributed} are proposed to deal with MTMCT. \cite{hofmann2013hypergraphs} constructs a mini-cost flow graph to complete data association among cameras in 3D world space. In \cite{berclaz2011multiple}, the data association is formulated as a constrained flow optimization of a convex problem, and the problem is solved by the k-shortest paths algorithm. %In \cite{wan2013distributed}, Wan \emph{et al.} model the global data association problem as searching the optimum labeling on a factor graph, and the framework can be implemented in a distributed manner.
In \cite{yoon2018multiple}, Yoon \emph{et al.} exploit the multiple hypothesis tracking (MHT) algorithm and apply it on MTMCT with some modifications. Branches in track-hypothesis trees represent the trajectory across multiple cameras. Maximum Weight Independent Set (MWIS) in \cite{papageorgiou2009maximum} is adopted for computing the best hypothesis set. With the development of Re-ID, a number of methods \cite{yoon2018multiple, liu2017multi, liu2017multi,zhang2017multi,ristani2018features} adopt Re-ID technology to represent the appearance of the target. In \cite{ristani2018features}, Ristani \emph{et al.} learn a good feature for both MTMCT and Re-ID with a convolutional neural network. In \cite{zhang2017multi}, Zhang \emph{et al.} obtain a good result with simple hierarchical clustering and well-trained Re-ID feature.

\subsection{Appearance feature}
In the context of appearance feature, many works \cite{ristani2018features,chu2019online,feng2019multi,zhu2018online,bae2018confidence,yoon2018online} recently adopt deep learning to represent appearance of the target. In \cite{feng2019multi}, Feng \emph{et al.} design a quality-aware mechanism to select the $K$ images from the historical samples of the target, and ResNet-18 \cite{he2016deep} is adopted to measure the quality of the detection. Then the Re-ID features of the selected detections are input into a classifier to get the similarity score between tracklets and detections. In \cite{zhu2018online}, spatial and temporal attention mechanism are adopted in feature extraction, which make the network focus on the matching patterns of the input image pair. In \cite{chu2017online}, Chu \emph{et al.} use spatial and temporal attention mechanism on feature extraction to handle the drift problem caused by single object tracker. In \cite{yoon2018online}, Yoon \emph{et al.} apply historical appearance matching to overcome the temporal error. The above methods attempt to solve the problems caused by occlusion and background clutter, and they maintain a stable appearance feature in a complex environment. In this paper, we employ the human pose information to estimate the target state including the occlusion status and orientation. In this way, we can make better use of Re-ID feature.
%-------------------------------------------------------------------------

\section{Proposed Method}

The overall design for MTMCT is introduced in this section. The proposed tracking framework consists of two parts: single camera tracking (SCT) and multiple camera tracking (MCT). In our work, the SCT tracker is utilized to generate trajectories in a single camera. Then a similar strategy as \cite{zhang2017multi} is adopted to cluster in-camera trajectories, and the final trajectories are obtained across multiple cameras.

The estimation of occlusion status and orientation are introduced in Sec \ref{sec:state_estimation}, and the fused tracking feature is described in Sec \ref{sec:fused_feat}. The overall SCT framework is presented in Sec \ref{sec:all_sct}. Finally, MCT tracker is presented in Sec \ref{sec:MCT}.

\subsection{State estimation}
\label{sec:state_estimation}
Occlusion status and orientation are estimated with the human pose information. Inference of the occlusion status and orientation is detailed as follows.
\vspace{-0.4cm}
\paragraph{Occlusion status estimation}
Human keypoints can be utilized to infer the occlusion status by the number of keypoints ($N_{valid}$) which are not occluded. And $N_{valid}$ is computed as:
\begin{equation}\label{equ:N_valid}
    N_{valid} = {\sum}_{i=1}^{N_k}\mathds{1}\{c_i > {\gamma}_{valid}\}
\end{equation}
where ${\gamma}_{valid}$ is the threshold for the confidence of keypoint $k_i$ to judge if $k_i$ is visible, $\mathds{1}$ equals 1 if the condition is true otherwise 0.

Re-ID feature is regarded as \textit{valid} when $N_{valid}$ is greater than the number threshold (${\theta}_{valid}$), which means that most of keypoints are visible and the target is not occluded, otherwise Re-ID feature is regarded as \textit{invalid}.

\begin{figure}[ht]
\centering
\includegraphics[scale=0.25]{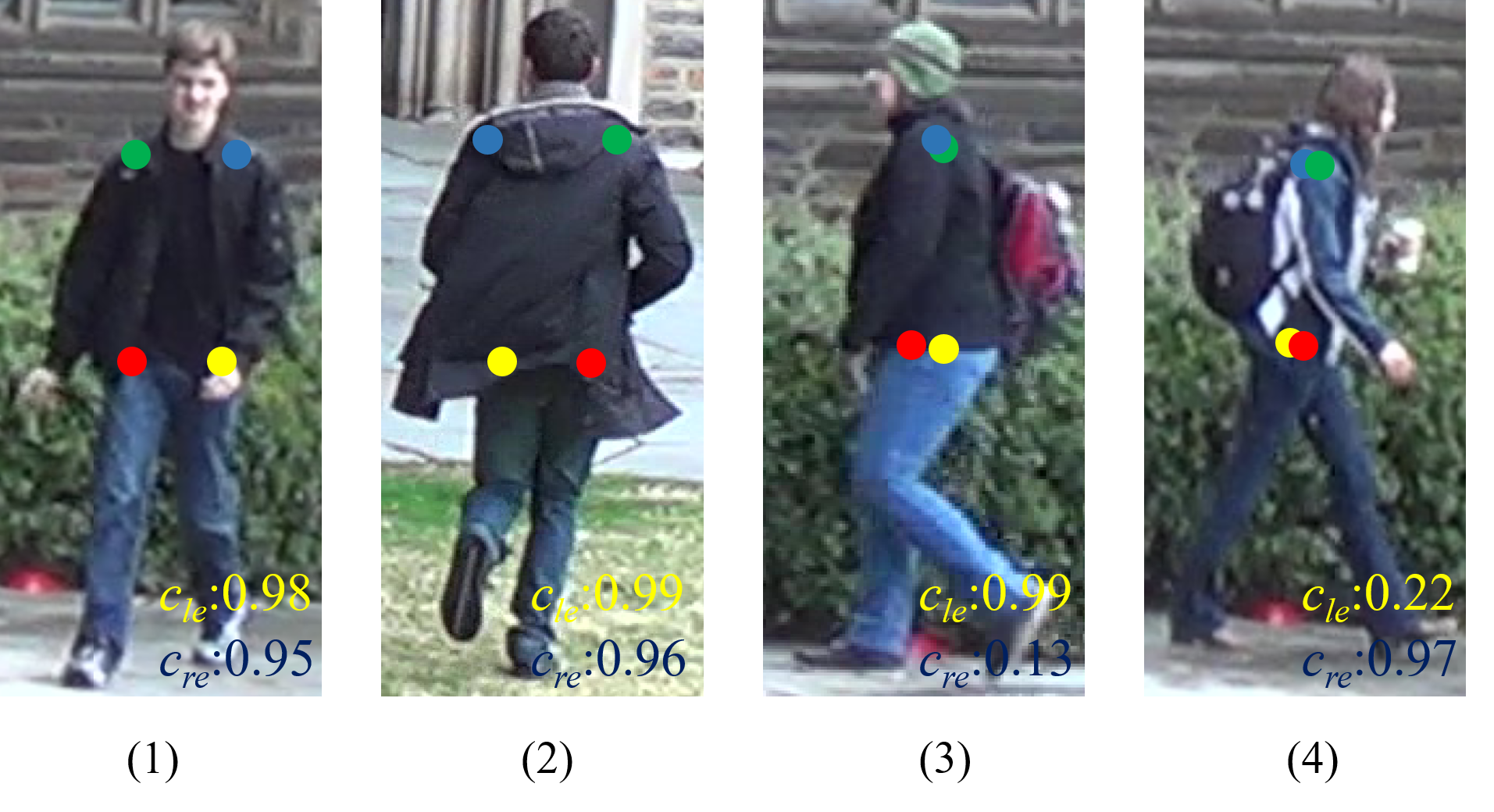}
\caption{4 orientations: (1) front, (2) back, (3) left and (4) right. The blue, green, red and yellow points represent the left shoulder, right shoulder, left hip and right hip keypoints respectively.}
\label{fig:orientation}
\end{figure}

\begin{figure*}[ht]
\centering
\includegraphics[scale=0.55]{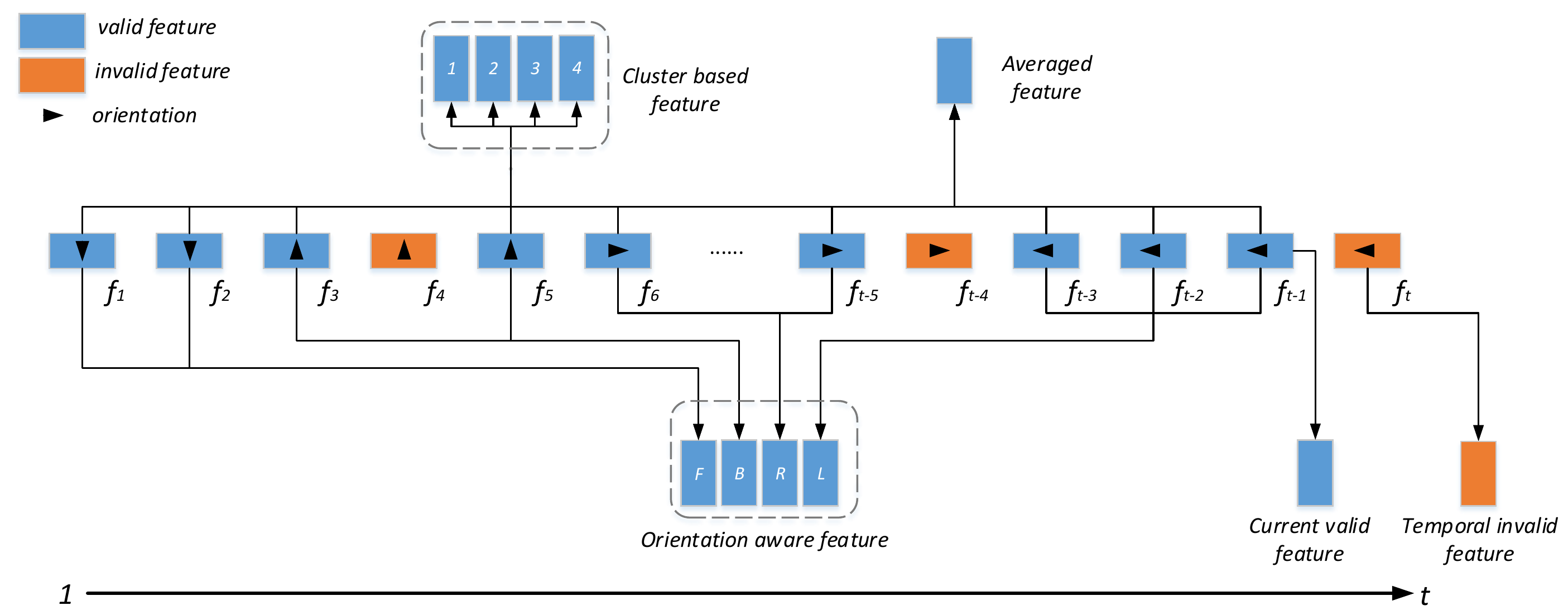}
\caption{The fused tracking feature consists of current valid feature, temporal invalid feature, orientation-aware feature and averaged feature. Saved historical Re-ID features of the tracklet is shown as $f_{1}, f_{2}, ..., f_{t-1}, f_{t}$, where $f_{i}$ is the Re-ID feature from the matched detection $d_{i}$ in frame $i$. Four types of orientations including front (F), back (B), left (L) and right (R) are illustrated in $f_{i}$. Cluster-based feature is shown with $N_{c}=4$. Detailed definition is presented in Sec \ref{sec:fused_feat}.}
\label{fig:fused_feature}
\end{figure*}

\begin{table}[!tp]
    \centering
    \caption{Architecture of the orientation classification network. A simple deep neural network is adopted to divide the input into four orientations. Five fully connected (FC) layers are utilized.}
    \label{tab:network}
    \begin{tabular}{ccc}
        \toprule
        Name& Input size& Output size\\
        \hline
        FC1   & 4 $\times$ 3 + 2 & 128 \\

        FC2   & 128 & 64 \\

        FC3   & 64 & 128 \\

        FC4   & 128 & 64 \\

        FC5   & 64 & 4 \\
        \bottomrule
    \end{tabular}

\end{table}
\vspace{-0.4cm}
\paragraph{Orientation estimation }
Orientation is an important cause for the appearance inconsistency of the same target. As illustrated in Fig. \ref{fig:orientation}, orientation can be easily estimated with body keypoint set $K_{body}=\{k_{ls}, k_{rs}, k_{lh}, k_{rh}\}$ corresponding to left shoulder, right shoulder, left hip, right hip and ear keypoint set $\{k_{le}, k_{re}\}$ corresponding to left ear, right ear.
In this work, the orientation is split into four states $O=\{o_{left}, o_{right}, o_{front}, o_{back}\}$, and orientation is inferred with Deep Neural Networks (DNNs) whose architecture is shown in Table \ref{tab:network}. Specifically, position and confidence of $K_{body}$, confidence of ears $c_{le}, c_{re}$ are input into DNN for classification task, so the input dimension is 14.

\subsection{Fused tracking feature}
\label{sec:fused_feat}
Due to the occlusion and variance of orientation, it is hard to model the appearance of the target with the growth of the tracklet. On the other hand, Re-ID model is widely adopted as an advanced appearance descriptor. However, most methods usually use Re-ID feature in a simple way like averaging these features. As shown in Fig. \ref{fig:fused_feature}, we adopt the well-designed fused tracking feature $F_{track}$ with many different combinations on saved historical Re-ID features of the tracklet, which can represent the appearance of the target more reliably.

The fused tracking feature $F_{track}$ of the tracklet is composed of five types of features as $F_{track}=\{f_{current}, f_{orientation}, f_{cluster}, f_{invalid}, f_{avg}\}$, which is illustrated in the following.
\vspace{-0.4cm}
\paragraph{Current valid feature} In some scenarios, the target moves fast so that their scale and pose change rapidly. We use the latest \textit{valid} feature in historical appearances of the target as $f_{current}$ to make the appearance model contain the latest information.
\vspace{-0.4cm}
\paragraph{Orientation-aware feature} Orientation-aware feature consists of four types of averaged features with different orientations $f_{orientation}=\{f_{left}, f_{right}, f_{front}, f_{back}\}$. Specifically, the feature of $f_{orientation}$ is the mean of all historical \textit{valid} features which have the same orientation.

In data association, one element in $f_{orientation}$ which shares the same orientation with the detection is chosen to compute the appearance distance.
Besides, The distance of $f_{orientation}$ between two tracklets is defined as the minimum among the Euclidean distances between the corresponding feature in the same orientation
\vspace{-0.4cm}
\paragraph{Cluster-based feature}
Feature clustering is widely used in non-supervised and semi-supervised Re-ID. In this work, an online cluster algorithm is employed on the cluster-based feature $f_{cluster}$ which has a similar initialization and updating strategy with the Gaussian mixture model. We set $N_c$ as the upper limit of the number of clusters in $f_{cluster}$ and $f_{cluster} = \varnothing$ as initialization. Algorithm \ref{alg:cluster_feature_update} details the updating strategy of $f_{cluster}$ when the tracklet matches a detection. % The center is the averaged feature in the cluster.

A $N_c \times N_c$ distance matrix $M_{cluster}$ is computed to obtain the distance $d_{cluster}$ between $f_{cluster}$ from two tracklets. Specifically, the value in $i$-$th$ row and $j$-$th$ column is the Euclidean distance between the $i$-$th$ cluster center and $j$-$th$ cluster center from two tracklets. The minimum value in $M_{cluster}$ is selected as $d_{cluster}$.

\begin{algorithm}
\caption{Updating $f_{cluster}$}
\label{alg:cluster_feature_update} % label of algorithm
\begin{algorithmic}[1] %
\REQUIRE Cluster-based feature is composed of $N$ clusters $f_{cluster} = \{C_1, C_2, ..., C_N\}$ and detections $d$ with Re-ID feature $f_d$%
\ENSURE updated $f_{cluster}$\\ %
% if-then-else\leftarrow
\IF{$f_d$ is \textit{invalid}}
\STATE return
\ELSE
\IF{$N<N_{c}$}
\STATE new cluster $C_{N+1}$ is initialized with $f_d$, $C_{N+1}=\{f_d\}$
\STATE $C_{N+1}$ is added to $f_{cluster}$
\ELSE
\STATE Get all cluster centers $f_{center}$ of $f_{cluster}$, $f_{center}=\{f_1, f_2, ..., f_N\}$
\FOR{$f_i$ of $f_{center}$}
\STATE $d_{ci}$ = $dist(f_i, f_d)$
\ENDFOR
\STATE $k=argmin\{d_{c1},d_{c2},..., d_{cN}\}$
\STATE The $k$-$th$ cluster $C_k$ is updated with $f_d$
\ENDIF
\ENDIF

\end{algorithmic}
\end{algorithm}
\vspace{-0.4cm}
\paragraph{Temporal invalid feature}
When the tracklet matches the detection with \textit{invalid} feature, the above three types of features do not update due to the unreliability of \textit{invalid} feature. However, IDS occurs if the appearance feature is not updated timely. Therefore, temporal invalid feature $f_{invalid}$ is adopted to update the \textit{invalid} feature and make the trajectory more smooth. It is worth noting that that $f_{invalid}$ only keeps the \textit{invalid} feature from the last frame, and will be removed from $F_{track}$ if expired.
\vspace{-0.4cm}
\paragraph{Averaged feature}
Averaged feature $f_{avg}$ is the feature averaged over all \textit{valid} Re-ID feature of the tracklet.

\label{subsec:fused_feature}

\begin{figure}[ht]
\centering
\includegraphics[scale=0.5]{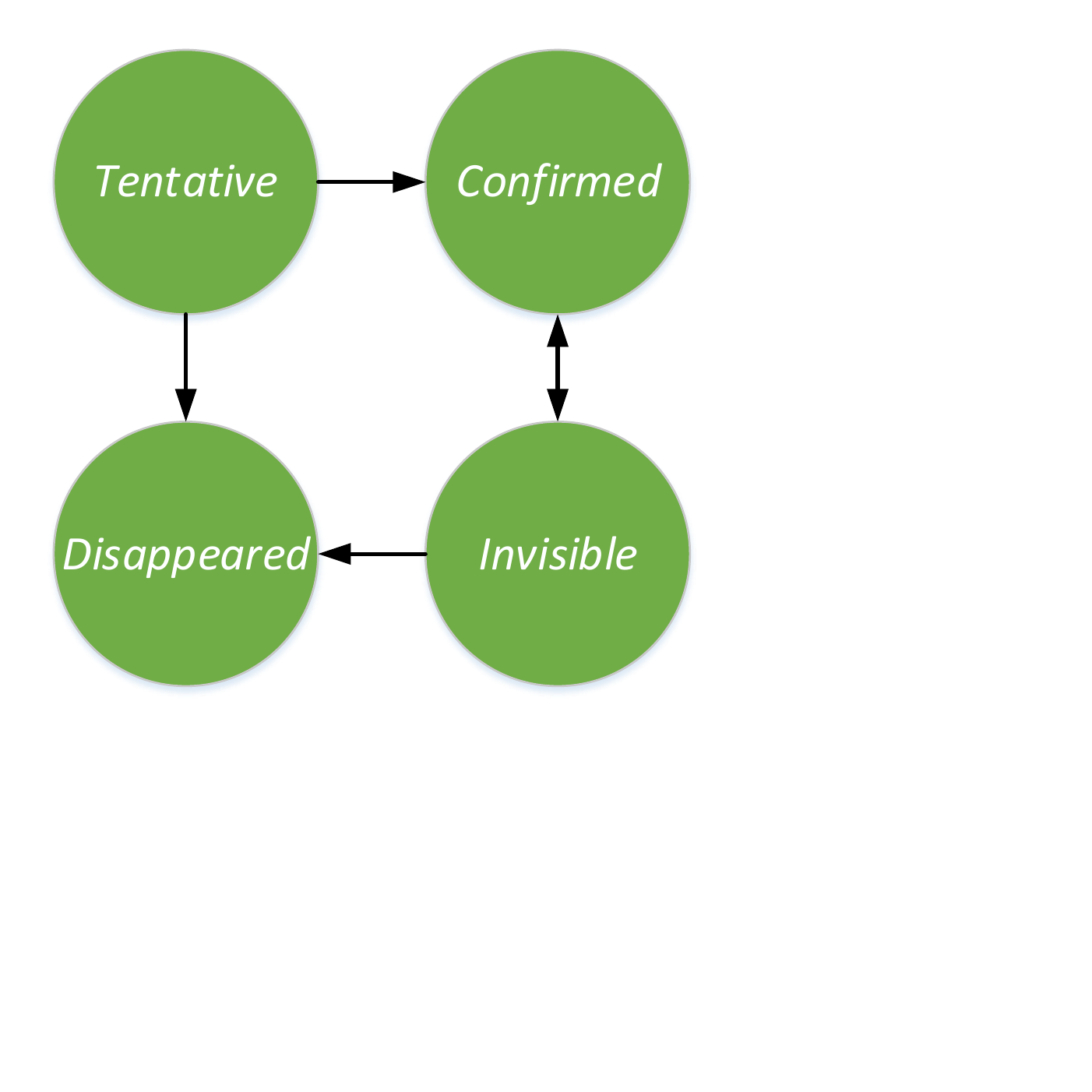}
\caption{Tracking phase and transformation of the tracklet. \textit{Tentative}, \textit{Confirmed}, \textit{Invisible} and \textit{Disappeared} are four phases of the tracklet lifetime. }
\label{fig:phase}
\end{figure}

\subsection{Single camera tracking}
\label{sec:all_sct}
\subsubsection{Tracking phase}
\label{sec:phase}
For modeling the lifetime of tracklet in the SCT tracker, we define four phases, \textit{Tentative}, \textit{Confirmed}, \textit{Invisible} and \textit{Disappeared} as shown in Fig. \ref{fig:phase}. New tracklet is generated with unmatched detection and initialized to different phase according to the occlusion status. If the detection is highly occluded, the tracking phase will be initialized as \textit{Tentative}. Otherwise, it will be initialized as \textit{Confirmed}. If tracklet in \textit{Confirmed} phase has been missed for ${\mu}_m$ times, it will enter \textit{Invisible} phase. And if tracklet in \textit{Invisible} phase has been missed for ${\mu}_d$ times, it will switch to \textit{Disappeared} phase. Tracklet in \textit{Invisible} phase will go back \textit{Confirmed} phase if the tracklet is matched in data association. Besides, tracklet in \textit{Tentative} phase will turn to \textit{Disappeared} phase if phase misses for one frame, and it will switch to \textit{Confirmed} phase if matches a detection with \textit{valid} feature. In this way, false positive detections can be removed. On the other hand, tracklet in \textit{Disappeared} phase means that the target is disappeared or has already left the scene, so the tracklet is removed from the tracklet set.

\begin{figure*}[ht]
\centering
\includegraphics[width=1.0\linewidth]{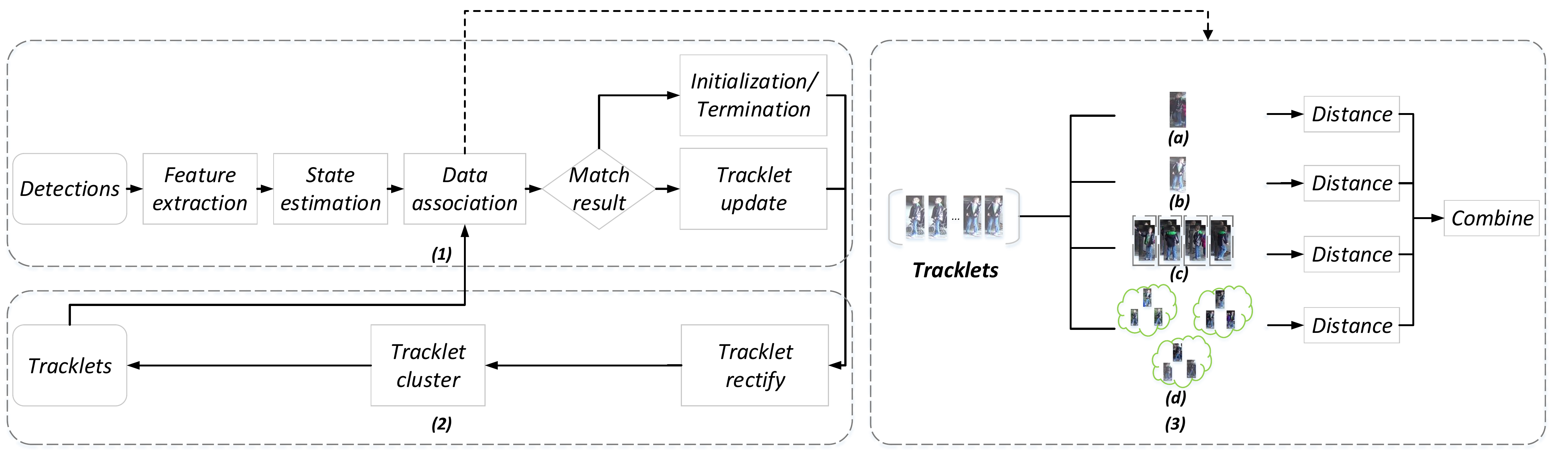}
   \caption{Proposed SCT framework. Overall tracking pipeline is shown in (1) and (2). Specifically, the pipeline of online tracklet generation is presented in (1), and the post-processing on existing tracklets is shown in (2). (3) details the computation of the distance matrix between existing tracklets and detections. (a), (b), (c), (d) are the temporal invalid feature, current valid feature, orientation-aware feature and cluster-based feature respectively. Distance is computed between Re-ID feature from detection and one of the fused tracking feature from tracklet. After that, the final distance between tracklet and detection is combined with the distances of four types of features.}
\label{fig:overall_design}
\end{figure*}

\subsubsection{Overall SCT framework}
\label{sec:sct}
In our SCT framework, tracking-by-detection strategy is adopted thanks to the development of object detection. Proposed tracking method follows the nearly online fashion to generate trajectory. Specifically, we maintain a tracklet set from the beginning to the end, and the tracking result is generated after tracklet clustering over every $K$ frames. The SCT framework in this paper can be divided into two parts: tracklets linking and tracklet association, which are shown in Fig. \ref{fig:overall_design} (1) and (2). (1) shows the pipeline which generates and updates tracklets in an online manner. (2) presents the post-processing on tracklets including tracklet rectifying and tracklet clustering. In this way, the tracklet fragment can be handled. Our online SCT framework is shown in detail as follows.

\begin{itemize}
    \vspace{-0.2cm}
    \item Step1. At current frame $t$, detections $D_t=\{d_i^t\}$ are obtained from the high-quality detector, Re-ID feature $f_i^t$ is extracted and keypoints $K_i^t$ of $d_i^t$ are estimated from the corresponding area.
    \vspace{-0.2cm}
    \item Step2. Estimate the occlusion status and orientation with the keypoints.
    \vspace{-0.2cm}
    \item Step3. Compute the Distance matrix $M_{dis} = \{d_{ij}\}$ between detections and maintained tracklet set.% $T=\{track_{j}\}$.
    \vspace{-0.2cm}
    \item Step4. Adopt Hungarian algorithm \cite{kuhn1955hungarian} on $M_{dis}$ to abtain the matching results including matched tracklets $T_{matched}$, unmatched tracklets $T_{unmatched}$ and unmatched detections $D_{unmatched}$.
    \vspace{-0.2cm}
    \item Step5. Update $T_{matched}$ with corresponding matched detections.
    \vspace{-0.2cm}
    \item Step6. Terminate the tracklets in $T_{unmatched}$ if disappear for a long time. Initialize new tracklets from $D_{unmatched}$ according to the occlusion status.
    \vspace{-0.2cm}
    \item Step7. Adopt the tracklet rectifying between tracklets in \textit{Confirmed} phase and tracklets in \textit{Invisible} phase, which are associated if satisfying the designed rule.
    \vspace{-0.2cm}
    \item Step8. Process the tracklet clustering for every $K$ frames in the current tracklet set, and generate the tracking result from the recent $K$ frames after clustering.

\end{itemize}

\subsubsection{Distance matrix in data association}
\label{sec:data_association}
Fig. \ref{fig:overall_design} (3) illustrates the distance between tracklet and detection and Algorithm \ref{affinity_calculate} describes the computation process. At first, elements in $M_{dis}$ are initialized to infinity value. Then we calculate distance between every pair $(t_{t-1}^i, d_t^i)$. Line 6 uses position information as the spatial constraint to prevent abnormal movement. Line 7 calculates the Euclidean distance between $f_{current}$ and $f_{det}$. Then the Euclidean distance $d_{orien}$ is calculated by the feature of $f_{orientation}$ which has the same orientation with $d_t^i$. In line 9, $d_{clu}$ is the minimum Euclidean distance between $f_{det}$ and every cluster center of $f_{cluster}$. In Line 10, $d_{invalid}$ will be computed if $f_{invalid}$ exists of $F_{track}$ and the occlusion status of the detection is \textit{invalid}. Finally the distance $m_i^j$ between $t_{t-1}^i$ and $d_{t}^j$ is set as the minimum value of $d_{curr},d_{orien}, d_{clu}$ and $d_{invalid}$.
% 待修改
\begin{algorithm}
\caption{Computing distance matrix for data association}
\label{affinity_calculate} % label of algorithm
\begin{algorithmic}[2] %
\REQUIRE tracklets $T_{t-1}$ in frame $t-1$ and detections $D_t$ in frame $t$ %
\ENSURE distance matrix $M_{dis} = \{m_i^j\}$\\ %
% if-then-else\leftarrow
\STATE initialization: $M_{dis}\leftarrow{inf}$
\FOR{each $t_{t-1}^i\in T_{t-1}$}
\FOR{each $d_t^j\in D_t$}
\STATE $d_{curr},d_{orien},d_{clu},d_{invalid}\longleftarrow inf$
\STATE get Re-ID feature $f_{det}$, human pose, orientation, occlusion status from $d_t^j$
\STATE check rationality using position information
\STATE $d_{curr}=dist(f_{current}, f_{det})$
\STATE $d_{orien}=dist(f_{orientation}, f_{det})$
\STATE $d_{clu} = dist(f_{cluster}, f_{det})$
\IF{$f_{invalid}$ exists and $f_{det}$ is \textit{invalid}}
\STATE $d_{invalid} = dist(f_{invalid}, f_{det})$
\STATE $m_i^j=min(d_{curr}, d_{orien}, d_{clu}, d_{invalid})$
\ELSE
\STATE $m_i^j=min(d_{curr}, d_{orien}, d_{clu})$
\ENDIF
\ENDFOR
\ENDFOR
\end{algorithmic}
\end{algorithm}

\vspace{-0.5cm}

\subsubsection{Tracklet association}
\label{sec:tracklet_association}
Tracklet association is crucial to linking the fragmented tracklets. The proposed tracker can re-track target after occlusion in two ways: tracklet rectifying and tracklet clustering. Before illustrating these two methods, we will introduce the physical constraints to prevent the impossible association and save the computation on constructing the distance matrix.
\paragraph{Physical constraints}
\label{phy_cons}
Three physical constraints are set to prevent the impossible association.
\begin{itemize}
    \item Tracklets can not be associated if they appear at the same time.
    \item Target can not move faster than a threshold. We analyze the position of ground truth and obtain the maximum possible velocity. Given two tracklets, we first sort them by time. The distance between the last detection of the former tracklet and the first detection from the latter one should be less than a maximum distance due to the constraint of velocity.
    \item Target can not disappear for a long time, so two tracklets can not be associated if the interval between two tracklets is larger than a threshold.
\end{itemize}
%% 有点难懂
\vspace{-0.4cm}
\paragraph{Tracklet rectifying} % 改为名词？
Tracklet rectifying is conducted on the \textit{Invisible} tracklets $T_{invisible}$ and \textit{Confirmed} tracklets $T_{confirmed}$, whose length has reached $L_{rectify}$. Tracklet rectifying aims to re-track the target after occlusion. When the target reappears after occlusion, it may not be matched by the previous tracklet at once, in which case a new tracklet is generated consequently. With the growth of new tracklet, $F_{track}$ becomes more stable, so we can use it to link the fragmented tracklets when the length of the newly generated tracklet reaches $L_{rectify}$. $f_{cluster}$ is adopted to measure the distance between $T_{invisible}$ and $T_{confirmed}$ in the distance matrix $M_{rectify}$ . The greedy algorithm is utilized on $M_{rectify}$ to obtain the matched pairs until the minimum distance over $\theta_{rectify}$. At last, the \textit{Invisible} tracklets is associated with corresponding \textit{Confirmed} tracklets according to the matching result.
\vspace{-0.4cm}
\paragraph{Tracklet clustering}
\label{sec:tracklet_cluster}
Tracklet clustering aims to associate all the tracklets except the \textit{Disappeared} ones. We follow the same strategy as tracklet rectifying. At first, distance matrix $M_{T-cluster}$ is constructed, then a greedy algorithm is adopted to associate the tracklets with the distance threshold $\theta_{cluster}$. While computing the distance matrix and associating the tracklets, same constraints are adopted. $f_{avg}$ and $f_{orientation}$ are utilized to compute the distance between two tracklets when constructing $M_{T-cluster}$. Specifically, distance between $f_{avg}$ and distance between $f_{orientation}$ are computed as $d_{avg}$ and $d_{ori}$, and the minor one between $d_{avg}$ and $d_{ori}$ is the final distance between two tracklets.

\subsection{Multi-camera tracking}
\label{sec:MCT}
Multi-camera tracking in this work is implemented with a distance matrix $M_{mct}$ and greedy algorithm, which is inspired by \cite{zhang2017multi}.

At first, we collect trajectories from all cameras and compute $M_{mct}$. We follow the same strategy to construct the distance matrix with \ref{sec:tracklet_cluster}.

After constructing $M_{mct}$, the greedy algorithm is adopted to associate trajectories until the minimum distance of the matrix exceeds $\theta_{mct}$. Besides, the same constraints are adopted as \cite{zhang2017multi} when associating trajectories. Different from \cite{zhang2017multi}, the distance is updated during associating. Specifically, when a trajectory is associated with others, the corresponding row and column are updated in $M_{mct}$. In addition, the SCT tracker in this paper is assumed to be good enough, and we only associate the trajectories across different cameras, so the in-camera trajectories stay unchanged.

\subsection{Implementation details}
\paragraph{Re-ID model}
We adopt ResNet-34 \cite{ResNet} to extract Re-ID feature. The size of input is $128 \times 256$ and 128-d Re-ID feature is extracted from the last fully connected layer. In training process, public Re-ID datasets are used including Market-1501 \cite{Market1501}, CUHK03 \cite{cuhk03}, MSMT17 \cite{msmt2017}, PRW \cite{PRW}, DukeMTMC-ReID \cite{zheng2017unlabeled} and extra private dataset. The Re-ID model achieves 78.5 Top1 accuracy and 62.4 mAP on DukeMTMC-ReID, whose performance is slightly worse than \cite{wu2017track}.
\vspace{-0.4cm}
\paragraph{Pose estimator}
In this paper, Alpha pose \cite{fang2017rmpe} is adopted to estimate the human pose. Besides, the pose estimator is not fine-tuned on the Duke MTMCT dataset.

\vspace{-0.4cm}
\paragraph{Parameter setting}
For modeling the lifetime of the target, $\mu_{m}$ is set as 10, and $\mu_{d}$ is set as 300. In state estimation, ${\gamma}_{valid}$ is set as 0.3, and $\theta_{valid}$ is set as 7. In the tracklet association, $\theta_{rectify}$, and $\theta_{cluster}$ are set as 20 and 30. In the multi-camera tracking, $\theta_{mct}$ is set as 40.

\section{Experiments}
In this section, experiments of the proposed state-aware MTMCT framework are conducted. First, the Duke MTMCT dataset and evaluation metric are introduced. Then the effectiveness of our work is proved, and we investigate the contribution of different components. Finally, the result on the test set is submitted, and the proposed tracking framework is compared with other state-of-the-art methods on this benchmark.

\begin{table*}[!tp]
\caption{The result of our tracker and several state-of-the-art trackers on test sequence of Duke MTMCT. The value in bold highlight is the best. Tracker MTMC\_{basel} is recently submitted on Duke MTMCT benchmark.}
\normalsize
  \centering
\begin{tabular}{|c|c|c|c|c|c|c|c|c|c|c|c|c|c|c|c|c|c|}
\hline

\multicolumn{1}{|c|}{ \multirow{2}*{Tracker} }& \multicolumn{3}{c|}{\textit{test-easy single} }& \multicolumn{3}{c|}{\textit{test-easy multiple} }& \multicolumn{3}{c|}{\textit{test-hard single} }& \multicolumn{3}{c|}{\textit{test-hard multiple}}\\
\cline{2-13}
\multicolumn{1}{|c|}{}&IDF1&IDP&IDR&IDF1&IDP&IDR&IDF1&IDP&IDR&IDF1&IDP&IDR\\
\hline
\multicolumn{1}{|c|}{BIPCC\cite{duke}}&70.1&83.6&60.4&56.2&67.0&48.4&64.5&81.2&53.5&47.3&59.6&39.2\\
\hline
\multicolumn{1}{|c|}{MYTRACKER\cite{yoon2018multiple}}&80.3&87.3&74.4&65.4&71.1&60.6&63.5&73.9&55.6&50.1&58.3&43.9\\
\hline
\multicolumn{1}{|c|}{TAREIDMTMC\cite{jiang2018online}}&83.8&87.6&80.4&68.8&71.8&66.0&77.9&86.6&70.7&61.2&68.0&55.5\\
\hline
\multicolumn{1}{|c|}{DeepCC\cite{ristani2018features}}&89.2&91.7&86.7&82.0&84.3&79.8&79.0&87.4&72.0&68.5&75.8&62.4\\
\hline
\multicolumn{1}{|c|}{MTMC\_ReID\cite{zhang2017multi}}&89.8&92.0&87.7&83.2&85.2&81.2&81.2&89.4&74.5&74.0&81.4&67.8\\
\hline
\multicolumn{1}{|c|}{MTMC\_{basel}}&91.3&91.8&\textbf{90.9}&\textbf{87.4}&87.8&\textbf{87.0}&83.7&88.8&79.1&75.4&80.0&71.3\\
\hline
\multicolumn{1}{|c|}{Ours}&\textbf{91.8}&\textbf{93.3}&90.3&86.8&\textbf{88.2}&85.4&\textbf{85.8}&\textbf{93.6}&\textbf{79.2}&\textbf{81.3}&\textbf{88.7}&\textbf{75.1}\\
\hline
\end{tabular}

\label{tab:compare}
\end{table*}

\subsection{Duke MTMCT dataset}
\label{sec:dataset}
The DukeMTMCT dataset is a large and detailed annotated dataset mainly for MTMCT task, which is recorded in Duke university with 8 cameras. There are 6,791 trajectories for 2,834 different identities overall and 25 minutes for each camera. The dataset is recorded at 60 FPS, and the resolution is 1080p. The dataset is split into three types of parts consisting of \textit{trainval}, \textit{test-easy}, \textit{test-hard}, and \textit{trainval-mini} is the subset of the \textit{trainval}, which contains 59281 frames.
\subsection{Evaluation metric}
In this work, ID Measure \cite{duke} is used as the criterion for both SCT and MTMCT tasks, which can measure the tracker performance globally. The performance evaluation is based on the truth-to-result match. Specifically, it constructs the matching matrix between ground truth and prediction trajectories and uses the Hungarian algorithm to get the final matching result. IDP, IDR and IDF1 are three main metrics for ID Measure. IDP (IDR) is the fraction of prediction (ground truth) detection are correctly identified. IDF1 is the correctly identified detections over the average value of ground truth and prediction.
\label{sec:everal_meptric}

\begin{table}[!tp]
    \centering
    \caption{Ablation study demonstrates the steady improvements of the state-aware Re-ID feature. IDF1, MOTA and IDS are shown in this table, and the arrows indicate low or high optimal metric values.}
    \label{tab:ablation_study}
    \begin{tabular}{|c|c|c|c|c|c|c|c|c|}
        \hline
        Method& IDF1 & MOTA & IDS \\
        \hline
        \makecell[cc]{Baseline}   & 77.1 & 82.8& 6409 \\
        \hline
        \makecell[cc]{Baseline + $f_{cluster}$}  & 82.5 & 82.6 & 8170 \\
        \hline
        \makecell[cc]{Baseline + $f_{cluster}$ + \\ $f_{orientation}$}  & 85.1 & 82.8 & 6564 \\
        \hline
        \makecell[cc]{Baseline + $f_{cluster}$ + \\ $f_{orientation}$ + $f_{invalid}$} & 85.2 &82.8 & 5466 \\
        \hline
    \end{tabular}

\end{table}

\subsection{Ablation study}
\label{sec:analysis}
Ablation study is conducted on the SCT task of \textit{trainval-mini} sequences. The orientation-aware feature, cluster-based feature and temporal valid feature are considered. The baseline tracker only uses current valid feature $f_{current}$ for data association and the time interval $K$ to generate tracking result is set as 10 seconds, which means 600 frames in Duke MTMCT dataset. The results of ablation study is shown in Table \ref{tab:ablation_study}.
\vspace{-0.5cm}
\paragraph{Cluster-based feature}
Comparing the tracker in the second row with baseline, the cluster-based feature is essential, and it can improve the performance on IDF1 by 5.4 \%. One can find that current valid feature can not model the target appearance correctly and the cluster-based feature is an effective way to represent the appearance of the target. $f_{cluster}$ becomes more stable with the growth of the tracklet, but is not robust to the occlusion at the beginning, which is the main cause for the increase of IDS.
\vspace{-0.5cm}
\paragraph{Orientation-aware feature}
Comparing the tracker in the third row with the tracker in the second row, the tracker with orientation-aware feature performs better which gains 2.6 \% improvement on IDF1. One can find that the orientation feature is complementary with the cluster feature.
\vspace{-0.5cm}
\paragraph{Temporal invalid Feature}
Comparing the tracker in the fourth row with the tracker in the third row, IDF1 is improved by 0.1 \%, and IDS is reduced from 6564 to 5466, which means the invalid temporal feature effectively reduces the IDS and makes the trajectory more smooth.

\subsection{Compare with other state-of-the-art methods}
\label{sec:compare}
We compare the proposed tracker with other tracking methods \cite{duke,yoon2018multiple,jiang2018online,ristani2018features,zhang2017multi} on DukeMTMCT dataset, and results are shown in Table \ref{tab:compare}.

We utilize the private detection provided by \cite{zhang2017multi}, and our tracker achieves a state-of-the-art performance on both \textit{test-easy} and \textit{test-hard} sequences. Besides, we outperform all officially published methods on IDF1 and IDR. We evaluate the proposed tracking method without any training or optimization on the train set, and the same parameters are utilized on \textit{test-easy} and \textit{test-hard} sequences. For the performance comparison, we collect some published methods and recently submitted method (MTMCT\_basel) on DukeMTMCT benchmark. For better performance, the offline SCT tracker is adopted, which means the time interval to cluster and output is set as the length of the corresponding sequence. Specifically, tracker first generates short but accurate tracklets and tracklet clustering in Sec. \ref{sec:tracklet_cluster} is adopted to associate these tracklets to obtain the final trajectories.

As shown in Table \ref{tab:compare}, our tracker achieves a new state-of-the-art performance on DukeMTMCT dataset. Due to the well-designed state-aware Re-ID feature, we outperform all other methods including an unpublished method on the benchmark by a large margin on \textit{test-hard}, which further verifies the robustness of proposed tracker in such crowded scene, as the motion cue becomes unstable when occluded.

\section{Conclusion}
In this paper, we propose the state-aware Re-ID feature for multiple cameras, multiple targets tracking task. We adopt human pose information to infer the occlusion status and orientation. Besides, the fused tracking feature is designed to make better use of Re-ID feature. Our tracker achieves a new state-of-the-art performance on DukeMTMCT benchmark, which verifies the effectiveness of the proposed method.

{\small
\bibliographystyle{ieee_fullname}
\bibliography{egbib}
}

\end{document}